\documentclass[]{article}
\usepackage{times,url,mathrsfs}
\usepackage{amsmath}
\usepackage{amsfonts}
\usepackage{graphicx}
\usepackage{subfigure}

\newtheorem{lemma}{Lemma}
\newtheorem{proof}{Proof}

\title{On Approximating the {\LARGE$l_p$} Distances for {\LARGE$p>2$} (When $p$ Is Even) }

\author{ {\bf Ping Li} \\
Department of Statistical Science \\
Faculty of Computing and Information Science\\
Cornell University, Ithaca,  NY 14850}

\begin{document}

\maketitle
\begin{abstract}

Many\footnote{First draft Dec. 2007. Slightly revised June 2008.} applications in machine learning and data mining require computing  pairwise $l_p$ distances in a data matrix $\mathbf{A}\in\mathbb{R}^{n\times D}$. For massive high-dimensional data, computing all pairwise distances of $\mathbf{A}$ can be infeasible. In fact, even  storing $\mathbf{A}$ or all pairwise distances of $\mathbf{A}$ in the memory may be also infeasible.

For $0<p\leq 2$, efficient small space algorithms exist, for example, based on the method of {\em stable random projections}, which unfortunately is not directly applicable to $p=3, 4, 5, 6, ...$ This paper proposes a simple method for $p = 2$, $4$, $6$, ... We first decompose the $l_p$ (where $p$ is even) distances into a sum of 2 marginal norms and $p-1$ ``inner products'' at different orders. Then we apply normal or sub-Gaussian random projections to approximate the resultant ``inner products,'' assuming that the marginal norms can be computed exactly by a linear scan.

We propose two strategies for applying random projections. The basic projection strategy requires only one projection matrix but it is more difficult to analyze, while the alternative projection strategy requires $p-1$ projection matrices but its theoretical analysis  is much easier. In terms of the accuracy, at least for $p=4$, the basic strategy is always more accurate than the alternative strategy if the data are non-negative, which is common in reality.

\end{abstract}



\section{Introduction}

This study proposes a simple method for efficiently computing the $l_p$ distances in a massive data matrix $\mathbf{A}\in\mathbb{R}^{n\times D}$ for $p>2$ (where $p$ is even), using {\em random projections}\cite{Book:Vempala}.

While many previous work on random projections focused on approximating the $l_2$ distances (and inner products), the  method of {\em symmetric stable random projections}\cite{Article:Indyk_TKDE03,Article:Indyk_JACM06,Article:Li_Hastie_Church_JMLR07,Proc:Li_SODA08} is applicable to approximating the $l_p$ distances for all $0<p\leq 2$. This work proposes using random projections for $p>2$, a least for some special cases.

Machine learning algorithms often operate on the $l_p$ distances of $\mathbf{A}$ instead of the original data. A straightforward application would be searching for the nearest neighbors using $l_p$ distance. The $l_p$ distance is also a basic loss functions for quality measure. The widely used ``kernel trick,'' (e.g., for support vector machines (SVM)), is often constructed on top of the $l_p$  distances\cite{Book:Scholkopf_02}.\footnote{It is well-known that the radial basis kernel using the $l_p$ distance with $0<p\leq 2$ satisfies the Mercer's condition. However, we can still use the $l_p$ distance with $p>2$ as kernels, although in this case  it is not guaranteed to find the ``most optimal'' solution. For very large-scale learning, we usually will not find the ``most optimal'' solution any way. }

Here we can treat $p$ as a {\em tuning} parameter.
It is common to take $p =2 $ ({\em Euclidian} distance), or $p =\infty$ ({\em infinity distance}), $p = 1$ ({\em Manhattan} distance),  or $p = 0$ ({\em Hamming} distance); but in principle any $p$ values are possible. In fact, if there is an efficient mechanism to compute the $l_p$ distances, then it becomes affordable to tune  learning algorithms for many values of $p$ for the best performance.

In modern data mining and learning applications, the ubiquitous phenomenon of ``massive data'' imposes challenges. For example, pre-computing and storing all pairwise $l_p$ distances in  memory at the cost $O(n^2)$ can be infeasible when $n>10^6$ (or even just $10^5$)\cite{Book:Bottou_07}. For ultra high-dimensional data, even just storing the whole data matrix can be infeasible. In the meanwhile, modern  applications can routinely involve millions of observations; and developing scalable  learning and data mining algorithms has  been an active research direction. One commonly used strategy in current practice is   to compute the distances \textbf{on the fly}\cite{Book:Bottou_07}, in stead of storing all pairwise $l_p$ distances.

Data reduction algorithms such as  sampling or sketching methods are also popular.
While there have been extensive studies on approximating the $l_p$ distances for $0<p\leq 2$, $p>2$ can be useful too. For example, because the normal distribution is completely determined by its first two moments (mean and variance), we can identify the non-normal components of the data by analyzing  higher moments, in particular, the fourth moments (i.e., {\em kurtosis}). Thus, the fourth moments are critical, for example, in the field of {\em Independent Component Analysis (ICA)}\cite{Book:ICA_01}. Therefore, it is viable to use the $l_p$ distance for $p>2$ when  lower order distances can not efficiently differentiate data.

It is unfortunate that the family of {\em stable distributions}\cite{Book:Zolotarev_86} is limited to $0<p\leq 2$ and hence we can not directly using {\em stable distributions} for approximating the $l_p$ distances. In the theoretical CS community, there have been many studies on approximating the $l_p$ norms and distances\cite{Proc:Alon_STOC96,Book:Henzinger_99,Proc:Feigenbaum_FOCS99,Proc:Indyk_FOCS00,Proc:Babcock_PODS02,Proc:Cormode_VLDB02,Article:Muthukrishnan_05,Proc:Saks_STOC02,Proc:Kumar_FOCS02,Proc:Woodruff_SODA04,Proc:Indyk_STOC05}, some of which also applicable to the $l_p$ distances (e.g., comparing two long vectors). Those papers proved that small space ($\hat{O}(1)$) algorithms exist only for $0<p\leq 2$.

\subsection{The Methodology}

Given a giant data matrix $\mathbf{A}\in\mathbb{R}^{n\times D}$, we assume that a linear scan of the data is feasible, but computing all pairwise interactions is not, either due to computational budget constraints or memory limits. Also, we only consider even $p = 4$, 6, ..., among which $p=4$ is probably the most important.

Interestingly, our method is based only on normal (or normal-like) projections. The observation is that, when $p$ is even, the $l_p$ distance can be decomposed into  marginal $l_p$ norms and ``inner products'' of various orders. For example, for two $D$-dimensional vectors $x$ and $y$, when $p = 4$, then
\begin{align}\notag
&d_{(p)} = \sum_{i=1}^D |x_i - y_i|^p
 = \sum_{i=1}^Dx_i^4 + \sum_{i=1}^Dy_i^4 + 6\sum_{i=1}^Dx_i^2y_i^2 - 4\sum_{i=1}^Dx_i^3y_i - 4\sum_{i=1}^Dx_iy_i^3.
\end{align}
Since we assume that a linear scan of the data is feasible, we can compute $\sum_{i=1}^Dx_i^4$ and $\sum_{i=1}^Dy_i^4$ exactly. We can approximate the interaction terms  $\sum_{i=1}^Dx_i^2y_i^2$,  $\sum_{i=1}^Dx_i^3y_i$, and  $\sum_{i=1}^Dx_iy_i^3$ using normal (or normal-like) random projections. Therefore, for $p$ being even, we are able to efficiently approximate the $l_p$ distances.

\subsection{Paper Organization}

Section \ref{sec_normal_4} concerns using normal random projections for approximating $l_4$ distances. We introduce two projection strategies and the concept of utilizing the marginal norms to improve the estimates.  Section \ref{sec_normal_6} extends this approach to approximating $l_6$ distances. Section \ref{sec_subG} analyzes the effect of replacing normal projections by sub-Gaussian projections.

\section{Normal Random Projections for $p=4$}\label{sec_normal_4}

The goal is to efficiently compute all pairwise $l_p$ ($p=4$) distances in $\mathbf{A}\in\mathbb{R}^{n\times D}$.
It suffices to consider any two rows of $\mathbf{A}$, say $x$ and $y$, where $x$, $y\in\mathbb{R}^{D}$. We need  to estimate the $l_p$ distance between $x$ and $y$
\begin{align}\notag
d_{(p)} = \sum_{i=1}^D|x_i - y_i|^p.
\end{align}
which, when $p = 4$, becomes
\begin{align}\notag
&d_{(4)} = \sum_{i=1}^D|x_i - y_i|^4
= \sum_{i=1}^D x_i^4 + \sum_{i=1}^D y_i^4
+ 6\sum_{i=1}^D x_i^2y_i^2 - 4\sum_{i=1}^D x_i^3y_i- 4\sum_{i=1}^D x_iy_i^3.
\end{align}

In one pass, we can compute $\sum_{i=1}^D x_i^4$ and $\sum_{i=1}^D y_i^4$ easily, but computing the interactions is more difficult. We resort to random projections for approximating $\sum_{i=1}^D x_i^2y_i^2$,  $\sum_{i=1}^D x_i^3y_i$, and $\sum_{i=1}^D x_iy_i^3$. Since there are three ``inner products'' of different orders, we can choose either only one projection matrix for all three terms (the basic projection strategy), or three independent projection matrices (the alternative projection strategy).

\subsection{The Basic Projection Strategy}

First, generate a random matrix $\mathbf{R}\in\mathbb{R}^{D\times k}$ ($k\ll D$), with i.i.d. entries\footnote{It is possible to relax the requirement of i.i.d samples. In fact, to prove unbiasedness of the estimates only needs pairwise independence, and to derive the variance formula requires four-wise independence.
}
 from a standard normal, i.e.,
\begin{align}\notag
&r_{ij} \sim N(0,1), \hspace{0.1in} \text{E}(r_{ij}) = 0, \hspace{0.1in} \text{E}(r_{ij}^2) = 1,  \hspace{0.1in} \text{E}(r_{ij}^4) = 3.\\\notag
&\text{E}\left(r_{ij}^sr_{i^\prime j^\prime}^t\right) = 0, \hspace{0.1in} \text{if}\  t \ \text{or}\  s \ \text{is odd, and } \ i\neq i^\prime \ \text{or} \ j\neq j^\prime
\end{align}

Using random projections, we generate six vectors in $k$ dimensions,  $u_1$, $u_2$, $u_3$, $v_1$, $v_2$, $v_3\in\mathbb{R}^{k}$:
\begin{align}\notag
&u_{1,j} = \sum_{i=1}^D x_i r_{ij}, \hspace{0.1in} u_{2,j} = \sum_{i=1}^D x_i^2 r_{ij}, \hspace{0.1in} u_{3,j} = \sum_{i=1}^D x_i^3 r_{ij},\\\notag
&v_{1,j} = \sum_{i=1}^D y_i r_{ij}, \hspace{0.1in} v_{2,j} = \sum_{i=1}^D y_i^2 r_{ij}, \hspace{0.1in} v_{3,j} =\sum_{i=1}^D y_i^3 r_{ij}.
\end{align}

We have a simple unbiased estimator of $d_{(4)}$
\begin{align}\notag
\hat{d}_{(4)} = \sum_{i=1}^D x_i^4 + \sum_{i=1}^D y_i^4 + \frac{1}{k}\left(6u_2^\text{T}v_2 - 4u_3^\text{T}v_1- 4u_1^\text{T}v_3\right).
\end{align}

\begin{lemma}\label{lem_var_4}
\begin{align}\notag
&\text{E}\left(\hat{d}_{(4)}\right) = d_{(4)},
\end{align}
\begin{align}\notag
\text{Var}\left( \hat{d}_{(4)}\right)
=&\frac{36}{k}\left(\sum_{i=1}^Dx_i^4\sum_{i=1}^Dy_i^4 +  \left(\sum_{i=1}^Dx_i^2y_i^2\right)^2\right)\\\notag
+& \frac{16}{k}\left(\sum_{i=1}^Dx_i^6\sum_{i=1}^Dy_i^2 +  \left(\sum_{i=1}^Dx_i^3y_i\right)^2\right)\\\notag
+&\frac{16}{k}\left(\sum_{i=1}^Dx_i^2\sum_{i=1}^Dy_i^6 +  \left(\sum_{i=1}^Dx_iy_i^3\right)^2\right)+\Delta_4
\end{align}
\begin{align}\notag
\Delta_4=-&\frac{48}{k}\left(\sum_{i=1}^D x_i^5\sum_{i=1}^Dy_i^3  +\sum_{i=1}^D x_i^2y_i \sum_{i=1}^Dx_i^3y_i^2\right)\\\notag
-&\frac{48}{k}\left(\sum_{i=1}^D x_i^3\sum_{i=1}^Dy_i^5 +\sum_{i=1}^D x_iy_i^2 \sum_{i=1}^Dx_i^2y_i^3\right)\\\notag
+&\frac{32}{k}\left(\sum_{i=1}^D x_i^4\sum_{i=1}^Dy_i^4 +\sum_{i=1}^D x_iy_i \sum_{i=1}^Dx_i^3y_i^3\right).
\end{align}
\begin{proof}
See Appendix \ref{proof_lem_var_4}. $\Box$
\end{proof}
\end{lemma}

The basic projection strategy is simple but its analysis is quite involved, especially when $p>4$. Also, if we are interested in higher order moments (other than variance) of the estimator, the analysis becomes very tedious.

\subsection{The Alternative Projection Strategy}\label{sec_alternative}

Instead of one projection matrix $\mathbf{R}$, we generate three, $\mathbf{R}^{(a)}$, $\mathbf{R}^{(b)}$, $\mathbf{R}^{(c)}$, independently.
By random projections, we generate six vectors in $k$ dimensions,  $u_1$, $u_2$, $u_3$, $v_1$, $v_2$, $v_3\in\mathbb{R}^{k}$, such that
\begin{align}\notag
&u_{1,j} = \sum_{i=1}^D x_i r_{ij}^{(c)}, \hspace{0.05in} u_{2,j} = \sum_{i=1}^D x_i^2 r_{ij}^{(a)}, \hspace{0.05in} u_{3,j} = \sum_{i=1}^D x_i^3 r_{ij}^{(b)},\\\notag
&v_{1,j} = \sum_{i=1}^D y_i r_{ij}^{(b)}, \hspace{0.05in} v_{2,j} = \sum_{i=1}^D y_i^2 r_{ij}^{(a)}, \hspace{0.05in} v_{3,j} =\sum_{i=1}^D y_i^3 r_{ij}^{(c)}.
\end{align}
Here we abuse the notation slightly by using the same $u$ and $v$ for both projection strategies.

Again, we have an unbiased estimator, denoted by $\hat{d}_{(4),a}$
\begin{align}\notag
\hat{d}_{(4),a} = \sum_{i=1}^D x_i^4 + \sum_{i=1}^D y_i^4 + \frac{1}{k}\left(6u_2^\text{T}v_2 - 4u_3^\text{T}v_1- 4u_1^\text{T}v_3\right)
\end{align}

\begin{lemma}
\begin{align}\notag
&\text{E}\left(\hat{d}_{(4),a}\right) = d_{(4)},
\end{align}
\begin{align}\notag
\text{Var}\left(\hat{d}_{(4),a}\right)
=&\frac{36}{k}\left(\sum_{i=1}^Dx_i^4\sum_{i=1}^Dy_i^4 +  \left(\sum_{i=1}^Dx_i^2y_i^2\right)^2\right)\\\notag
+& \frac{16}{k}\left(\sum_{i=1}^Dx_i^6\sum_{i=1}^Dy_i^2 +  \left(\sum_{i=1}^Dx_i^3y_i\right)^2\right)\\\notag
+&\frac{16}{k}\left(\sum_{i=1}^Dx_i^2\sum_{i=1}^Dy_i^6 +  \left(\sum_{i=1}^Dx_iy_i^3\right)^2\right).
\end{align}
\begin{proof}
The proof basically follows from that of Lemma \ref{lem_var_4}.
\end{proof}
\end{lemma}

Compared with $\text{Var}\left(\hat{d}_{(4)}\right)$ in Lemma \ref{lem_var_4}, the difference would be $\Delta_4$
\begin{align}\notag
&\text{Var}\left(\hat{d}_{(4)}\right)-\text{Var}\left(\hat{d}_{(4),a}\right) = \Delta_4\\\notag
=&-\frac{48}{k}\left(\sum_{i=1}^D x_i^5\sum_{i=1}^Dy_i^3  +\sum_{i=1}^D x_i^2y_i \sum_{i=1}^Dx_i^3y_i^2\right)\\\notag
&-\frac{48}{k}\left(\sum_{i=1}^D x_i^3\sum_{i=1}^Dy_i^5 +\sum_{i=1}^D x_iy_i^2 \sum_{i=1}^Dx_i^2y_i^3\right)\\\label{eqn_var_diff1}
&+\frac{32}{k}\left(\sum_{i=1}^D x_i^4\sum_{i=1}^Dy_i^4 +\sum_{i=1}^D x_iy_i \sum_{i=1}^Dx_i^3y_i^3\right),
\end{align}
which can be either negative or positive. For example, when all $x_i$'s are negative and all $y_i$'s are positive, then $\Delta_4\geq0$, i.e.,  the alternative projections strategy results in smaller variance and hence it should be adopted.

We can show in Lemma \ref{lem_var_diff} that when the data are non-negative (which is more likely the reality),  the difference in (\ref{eqn_var_diff1}) will never exceed zero, suggesting that the basic strategy would be preferable, which is also operationally simpler (although more sophisticated in the analysis).

\begin{lemma}\label{lem_var_diff}
If all entries of $x$ and $y$ are non-negative, then
\begin{align}\label{eqn_var_diff}
\text{Var}\left(\hat{d}_{(4)}\right)-\text{Var}\left(\hat{d}_{(4),a}\right) = \Delta_4\leq 0.
\end{align}

\begin{proof}

See Appendix \ref{proof_lem_var_diff}. $\Box$.

\end{proof}
\end{lemma}

Thus, the main advantage of the alternative projection strategy is that it simplifies the analysis, especially true when $p>4$. Also, analyzing the alternative projection strategy may provide an estimate for the basic projection strategy. For example, the variance of $\hat{d}_{(4),a}$ is an upper bound of the variance of $\hat{d}_{(4)}$ in non-negative data.

In the next subsection, we show that the alternative strategy make the analysis feasible when we take advantage of the marginal information.

\subsection{Improving the Estimates Using Margins}

Since we assume that a linear scan of the data is feasible and in fact the estimators in both strategies already take advantage of the marginal $l_4$ norms, $\sum_{i=1}^Dx_i^4$ and $\sum_{i=1}^Dy_i^4$, we might as well compute other marginal norms and try to take advantage of them in a systematic manner.

Lemma \ref{lem_margin} demonstrates such a method for improving  estimates using margins. For simplicity, we assume in Lemma \ref{lem_margin} that we adopt the alternative projection strategy, in order to carry out the (asymptotic) analysis of the variance.

\begin{lemma}\label{lem_margin}
Suppose we use the alternative projection strategy described in Section \ref{sec_alternative} to generate samples $u_{1,j}$, $u_{2,j}$, $u_{3,j}$, $v_{1,j}$, $v_{2,j}$, and $v_{3,j}$.  We estimate $d_{(4)}$ by
\begin{align}\notag
\hat{d}_{(4),a,mle} = \sum_{i=1}^Dx_i^4+\sum_{i=1}^Dy_i^4 + 6\hat{a}_{2,2} - 4\hat{a}_{3,1} - 4\hat{a}_{1,3},
\end{align}
where $\hat{a}_{2,2}$,  $\hat{a}_{3,1}$, $\hat{a}_{1,3}$, are respectively, the solutions to the following three cubic equations:
{\small\begin{align}\notag
&a_{2,2}^3 -\frac{a_{2,2}^2}{k}u_2^\text{T}v_2 - \frac{1}{k}\sum_{i=1}^Dx_i^4\sum_{i=1}^Dy_i^4 u_2^\text{T}v_2 \\\notag
&\hspace{0.in}+ {a_{2,2}}\left(-\sum_{i=1}^Dx_i^4\sum_{i=1}^Dy_i^4\right)+\frac{a_{2,2}}{k}\left(\sum_{i=1}^Dx_i^4\|v_2\|^2 + \sum_{i=1}^Dy_i^4\|u_2\|^2\right) = 0.
\end{align}
\begin{align}\notag
&a_{3,1}^3 -\frac{a_{3,1}^2}{k}u_3^\text{T}v_1 - \frac{1}{k}\sum_{i=1}^Dx_i^6\sum_{i=1}^Dy_i^2 u_3^\text{T}v_1 \\\notag
&\hspace{0.in}+ {a_{3,1}}\left(-\sum_{i=1}^Dx_i^6\sum_{i=1}^Dy_i^2\right)+\frac{a_{3,1}}{k}\left(\sum_{i=1}^Dx_i^6\|v_1\|^2 + \sum_{i=1}^Dy_i^2\|u_3\|^2\right) = 0.
\end{align}
\begin{align}\notag
&a_{1,3}^3 -\frac{a_{1,3}^2}{k}u_1^\text{T}v_3 - \frac{1}{k}\sum_{i=1}^Dx_i^2\sum_{i=1}^Dy_i^6 u_1^\text{T}v_3 \\\notag
&\hspace{0.in}+ {a_{1,3}}\left(-\sum_{i=1}^Dx_i^2\sum_{i=1}^Dy_i^6\right)+\frac{a_{1,3}}{k}\left(\sum_{i=1}^Dx_i^2\|v_3\|^2 + \sum_{i=1}^Dy_i^6\|u_1\|^2\right) = 0.
\end{align}}

Asymptotically (as $k\rightarrow\infty$), the variance  would be
\begin{align}\notag
&\text{Var}\left(\hat{d}_{(4),a,mle}\right) \\\notag
=& 36\text{Var}\left(\hat{a}_{2,2}\right) + 16\text{Var}\left(\hat{a}_{2,2}\right)+ 16\text{Var}\left(\hat{a}_{2,2}\right)\\\notag
=& \frac{36}{k} \frac{\left(\sum_{i=1}^Dx_i^4\sum_{i=1}^Dy_i^4 - \left(\sum_{i=1}^Dx_i^2y_i^2\right)^2\right)^2}{
\sum_{i=1}^Dx_i^4\sum_{i=1}^Dy_i^4 + \left(\sum_{i=1}^Dx_i^2y_i^2\right)^2
}\\\notag
+& \frac{16}{k} \frac{\left(\sum_{i=1}^Dx_i^6\sum_{i=1}^Dy_i^2 - \left(\sum_{i=1}^Dx_i^3y_i\right)^2\right)^2}{
\sum_{i=1}^Dx_i^6\sum_{i=1}^Dy_i^2 + \left(\sum_{i=1}^Dx_i^3y_i\right)^2
}\\\notag
+& \frac{16}{k} \frac{\left(\sum_{i=1}^Dx_i^2\sum_{i=1}^Dy_i^6 - \left(\sum_{i=1}^Dx_iy_i^3\right)^2\right)^2}{
\sum_{i=1}^Dx_i^2\sum_{i=1}^Dy_i^6 + \left(\sum_{i=1}^Dx_iy_i^3\right)^2
}
+O\left(\frac{1}{k^2}\right)
\end{align}
\begin{proof}
 \cite{Proc:Li_Hastie_Church_COLT06,Proc:Li_Hastie_Church_KDD06} proposed taking advantage of the marginal $l_2$ norms to improve the estimates of $l_2$ distances and inner products. Because we assume the alternative projection strategy, we can analyze $\hat{a}_{2,2}$, $\hat{a}_{3,1}$, and $\hat{a}_{1,3}$, independently and then combine the results; and hence we skip the detailed proof.
\end{proof}
\end{lemma}

Of course, in practice, we probably still prefer the basic projection strategy, i.e., only one projection matrix instead of three. In this case, we still solve three cubic equations, but the precise analysis of the variance becomes much more difficult. When the data are non-negative, we believe that $\text{Var}\left(\hat{d}_{(4),a,mle}\right)$ will also be the upper bound of the estimation variance using the basic projection strategy, which can be easily verified by empirical results (not included in the current report).

Solving cubic equations is  easy, as there are closed-form solutions. We can also solve the equations by iterative methods. In fact, it is common practice to do only a one-step iteration (starting with the solution without using margins), called ``one-step Newton-Rhapson'' in statistics.

\section{Normal Random Projections for P=6}\label{sec_normal_6}

For higher $p$ (where $p$ is even), we can follow basically the same procedure as for $p=4$. To illustrate this, we work out an example for $p=6$. We only demonstrate the basic projection strategy.

The $l_6$ distance can be decomposed into 2 marginal norms and 5 inner products at various orders:
\begin{align}\notag
&{d}_{(6)} = \sum_{i=1}^D x_i^6 + \sum_{i=1}^D y_i^6 - 20\sum_{i=1}^Dx_i^3y_i^3\\\notag
&+15\sum_{i=1}^Dx_i^2y_i^4 + 15\sum_{i=1}^Dx_i^4y_i^2 - 6\sum_{i=1}^Dx_i^5y_i - 6\sum_{i=1}^Dx_iy_i^5
\end{align}

Generate one random projection matrix $\mathbf{R}\in\mathbb{R}^{D\times k}$, and
\begin{align}\notag
&u_{1,j} = \sum_{i=1}^D x_i r_{ij}, \hspace{0.in} u_{2,j} = \sum_{i=1}^D x_i^2 r_{ij}, \hspace{0.in} u_{3,j} = \sum_{i=1}^D x_i^3 r_{ij},\\\notag
&\hspace{0.in} u_{4,j} = \sum_{i=1}^D x_i^4 r_{ij}, \hspace{0.in} u_{5,j} = \sum_{i=1}^D x_i^5 r_{ij},\\\notag
&v_{1,j} = \sum_{i=1}^D y_i r_{ij}, \hspace{0.in} v_{2,j} = \sum_{i=1}^D y_i^2 r_{ij}, \hspace{0.in} v_{3,j} =\sum_{i=1}^D y_i^3 r_{ij},\\\notag
&v_{4,j} = \sum_{i=1}^D y_i^4 r_{ij}, \hspace{0.in} v_{5,j} =\sum_{i=1}^D y_i^5 r_{ij}.
\end{align}

Lemma \ref{lem_var_6} provide the variance of the following unbiased estimator of $d_{(6)}$:
{\small\begin{align}\notag
&\hat{d}_{(6)} = \sum_{i=1}^D x_i^6 + \sum_{i=1}^D y_i^6+ \frac{1}{k}\left(-20u_3^\text{T}v_3 + 15u_4^\text{T}v_2+ 15u_2^\text{T}v_4-6u_5^\text{T}v_3- 6u_1^\text{T}v_5\right)\\\notag
&=\sum_{i=1}^D x_i^6 + \sum_{i=1}^D y_i^6 + \frac{1}{k}
\sum_{j=1}^k -20u_{3,j}v_{3,j} + 15u_{2,j} v_{4,j} + 15u_{4,j} v_{2,j} - 6u_{1,j} v_{5,j} - 6u_{5,j} v_{1,j}.
\end{align}}

\begin{lemma}\label{lem_var_6}
{\small\begin{align}\notag
\text{Var}\left( \hat{d}_{(6)}\right)
&=\frac{400}{k}\left(\sum_{i=1}^Dx_i^6\sum_{i=1}^Dy_i^6 +  \left(\sum_{i=1}^Dx_i^3y_i^3\right)^2\right)
+\frac{225}{k}\left(\sum_{i=1}^Dx_i^4\sum_{i=1}^Dy_i^8 +  \left(\sum_{i=1}^Dx_i^2y_i^4\right)^2\right)\\\notag
&+\frac{225}{k}\left(\sum_{i=1}^Dx_i^8\sum_{i=1}^Dy_i^4 +  \left(\sum_{i=1}^Dx_i^4y_i^2\right)^2\right)
+\frac{36}{k}\left(\sum_{i=1}^Dx_i^2\sum_{i=1}^Dy_i^{10} +  \left(\sum_{i=1}^Dx_iy_i^5\right)^2\right)\\\notag
&+\frac{36}{k}\left(\sum_{i=1}^Dx_i^{10}\sum_{i=1}^Dy_i^2 +  \left(\sum_{i=1}^Dx_i^5y_i\right)^2\right)+\Delta_6
\end{align}}
where
{\small\begin{align}\notag
\Delta_6=&
-\frac{600}{k}\left(\sum_{i=1}^D x_i^5\sum_{i=1}^Dy_i^7 + \sum_{i=1}^D x_i^3y_i^4 \sum_{i=1}^Dx_i^2y_i^3\right)
-\frac{600}{k}\left(\sum_{i=1}^D x_i^7\sum_{i=1}^Dy_i^5 +\sum_{i=1}^D x_i^3y_i^2 \sum_{i=1}^Dx_i^4y_i^3\right)\\\notag
&+\frac{240}{k}\left(\sum_{i=1}^D x_i^4\sum_{i=1}^Dy_i^8 +\sum_{i=1}^D x_i^3y_i^5 \sum_{i=1}^Dx_iy_i^3\right)
+\frac{240}{k}\left(\sum_{i=1}^D x_i^8\sum_{i=1}^Dy_i^4 + \sum_{i=1}^D x_i^3y_i \sum_{i=1}^Dx_i^5y_i^3\right)\\\notag
&+\frac{450}{k}\left(\sum_{i=1}^D x_i^6\sum_{i=1}^Dy_i^6 +\sum_{i=1}^D x_i^2y_i^2 \sum_{i=1}^Dx_i^4y_i^4\right)
-\frac{180}{k}\left(\sum_{i=1}^D x_i^3\sum_{i=1}^Dy_i^9 +\sum_{i=1}^D x_i^2y_i^5 \sum_{i=1}^Dx_iy_i^4\right)\\\notag
&-\frac{180}{k}\left(\sum_{i=1}^D x_i^7\sum_{i=1}^Dy_i^5 +\sum_{i=1}^D x_i^2y_i \sum_{i=1}^Dx_i^5y_i^4\right)
-\frac{180}{k}\left(\sum_{i=1}^D x_i^5\sum_{i=1}^Dy_i^7 +\sum_{i=1}^D x_i^4y_i^5 \sum_{i=1}^Dx_iy_i^2\right)\\\notag
&-\frac{180}{k}\left(\sum_{i=1}^D x_i^9\sum_{i=1}^Dy_i^3 +\sum_{i=1}^D x_i^4y_i \sum_{i=1}^Dx_i^5y_i^2\right)
+\frac{72}{k}\left(\sum_{i=1}^D x_i^6\sum_{i=1}^Dy_i^6 +\sum_{i=1}^D x_iy_i \sum_{i=1}^Dx_i^5y_i^5\right).
\end{align}}
\begin{proof}
See Appendix \ref{proof_lem_var_6}. $\Box$.
\end{proof}
\end{lemma}

When all entries of $x$ and $y$ are non-negative, we believe it is true that $\Delta_6 \leq 0$, but we did not proceed with the proof.

Of course, it is again a good idea to take advantage of the marginal norms, but we skip the analysis.

\section{Sub-Gaussian Random Projections}\label{sec_subG}
It is well-known that it is not necessary to sample {\small$r_{ij}\sim N(0,1)$}. In fact, to have an unbiased estimator, it suffices to sample $r_{ij}$ from any distribution with zero mean (and unit variance). For good higher-order behaviors, it is often a good idea to sample from a {\em sub-Gaussian} distribution, of which a zero-mean normal distribution is a special case.

The theory of {\em sub-Gaussian} distributions was developed in the 1950's. See \cite{Book:Buldygin_00} and references therein.
A random variable $x$ is  {\em sub-Gaussian}  if there exists a
constant $g>0$ such that for  all $t\in \mathbb{R}$:
\begin{align}\notag
\text{E}\left(\exp(xt)\right) \leq \exp\left(\frac{g^2t^2}{2}\right).
\end{align}

In this section, we sample $r_{ij}$ from a sub-Gaussian distribution with the following restrictions:
\begin{align}\notag
\text{E}(r_{ij})  = 0, \hspace{0.2in} \text{E}(r_{ij})  = 1, \hspace{0.2in} \text{E}(r_{ij}^4)  = s,
\end{align}
and we denote $r_{ij}\sim SubG(s)$.  It can be shown that we must restrict $s\geq 1$.

One example would be the $r_{ij}\sim Uniform (-\sqrt{3}, \sqrt{3})$, for which $s = \frac{9}{5}$. Although the uniform distribution is simpler than normal, it is now well-known that we should sample from the following three-point sub-Gaussian distributions\cite{Proc:Achlioptas_PODS01}.
\begin{align}\notag
r_{ij} = \sqrt{s}\times\left\{\begin{array}{rl} 1 & \text{ with prob. }
    \frac{1}{2s}  \\ 0 & \text{ with prob. } 1-\frac{1}{s}\\ -1 & \text{ with prob. }
    \frac{1}{2s} \end{array} \right.,\hspace{0.3in} s\geq 1
\end{align}

In our analysis, we do not have to specify the exact distribution of $r_{ij}$ and we can simply express the estimation variance as a function  of $s$.

Here, we consider the basic projections strategy, by generating one random projection matrix $\mathbf{R}\in\mathbb{R}^{n\times D}$ with i.i.d. entries $r_{ij}\sim SubG(s)$, and
\begin{align}\notag
&u_{1,j} = \sum_{i=1}^D x_i r_{ij}, \hspace{0.in} u_{2,j} = \sum_{i=1}^D x_i^2 r_{ij}, \hspace{0.in} u_{3,j} = \sum_{i=1}^D x_i^3 r_{ij},\\\notag
&v_{1,j} = \sum_{i=1}^D y_i r_{ij}, \hspace{0.in} v_{2,j} = \sum_{i=1}^D y_i^2 r_{ij}, \hspace{0.in} v_{3,j} =\sum_{i=1}^D y_i^3 r_{ij}.
\end{align}

We again have a simple unbiased estimator of $d_{(4)}$
\begin{align}\notag
\hat{d}_{(4),s} = \sum_{i=1}^D x_i^4 + \sum_{i=1}^D y_i^4 + \frac{1}{k}\left(6u_2^\text{T}v_2 - 4u_3^\text{T}v_1- 4u_1^\text{T}v_3\right)
\end{align}

\begin{lemma}\label{proof_lem_var_subG}
{\small\begin{align}\notag
\text{Var}\left(\hat{d}_{(4),s}\right)
=&\frac{36}{k}\left(\sum_{i=1}^Dx_i^4\sum_{i=1}^Dy_i^4 +  \left(\sum_{i=1}^Dx_i^2y_i^2\right)^2+(s-3)\sum_{i=1}^D x_i^4y_i^4\right)\\\notag
+& \frac{16}{k}\left(\sum_{i=1}^Dx_i^6\sum_{i=1}^Dy_i^2 +  \left(\sum_{i=1}^Dx_i^3y_i\right)^2+(s-3)\sum_{i=1}^D x_i^6y_i^2\right)\\\notag
+&\frac{16}{k}\left(\sum_{i=1}^Dx_i^2\sum_{i=1}^Dy_i^6 +  \left(\sum_{i=1}^Dx_iy_i^3\right)^2+(s-3)\sum_{i=1}^D x_i^2y_i^6\right)\\\notag
-&\frac{48}{k}\left(\sum_{i=1}^D x_i^5\sum_{i=1}^Dy_i^3  +\sum_{i=1}^D x_i^2y_i \sum_{i=1}^Dx_i^3y_i^2+(s-3)\sum_{i=1}^D x_i^5y_i^3\right)\\\notag
-&\frac{48}{k}\left(\sum_{i=1}^D x_i^3\sum_{i=1}^Dy_i^5 +\sum_{i=1}^D x_iy_i^2 \sum_{i=1}^Dx_i^2y_i^3+(s-3)\sum_{i=1}^D x_i^3y_i^5\right)\\\notag
+&\frac{32}{k}\left(\sum_{i=1}^D x_i^4\sum_{i=1}^Dy_i^4 +\sum_{i=1}^D x_iy_i \sum_{i=1}^Dx_i^3y_i^3+(s-3)\sum_{i=1}^D x_i^4y_i^4\right).
\end{align}}

\begin{proof}
See Appendix \ref{proof_lem_var_subG}. $\Box$.
\end{proof}
\end{lemma}

\section{Conclusions}

It has been an active research topic on approximating  $l_p$ distances in massive high-dimensional data, for example, a giant ``data matrix'' $\mathbf{A}\in\mathbb{R}^{n\times D}$. While a linear scan on $\mathbf{A}$ may be feasible, it can be prohibitive (or even infeasible) to compute and store all pairwise $l_p$ distances. Using random projections can reduce the cost of computing all pairwise distances from $O(n^2D)$ to $(n^2k)$ where $k\ll D$. The data size is reduced from $O(nD)$ to $O(nk)$ and hence it may be possible to store the reduced data in memory.

While the well-known method of {\em stable random projections} is applicable to $0<p\leq 2$, not directly to $p>2$,   we propose a practical approach for approximating the $l_p$ distances in massive data for $p=2, 4, 6, ...$, based on the simple fact that, when $p$ is even, the $l_p$ distances can be decomposed into 2 marginal norms and $p-1$ ``inner products'' of various orders. Two projection strategies are proposed to approximate these ``inner products'' as well as the $l_p$ distances; and we show the basic projection strategy (which is simpler) is always preferable over the alternative strategy in terms of the accuracy, at least for $p = 4$ in non-negative data. We also propose utilizing the marginal norms (which can be easily computed exactly) to further improve the estimates. Finally, we  analyze the performance using sub-Gaussian random projections.

{\small

\appendix
\section{Proof of Lemma 1}\label{proof_lem_var_4}
{\small\begin{align}\notag
&\hat{d}_{(4)} = \sum_{i=1}^D x_i^4 + \sum_{i=1}^D y_i^4 + \frac{1}{k}\left(6u_2^\text{T}v_2 - 4u_3^\text{T}v_1- 4u_1^\text{T}v_3\right)\\\notag
=&\sum_{i=1}^D x_i^4 + \sum_{i=1}^D y_i^4 + \frac{1}{k}\left(
\sum_{j=1}^k 6u_{2,j}v_{2,j} - 4u_{3,j} v_{1,j} - 4u_{1,j} v_{3,j}
\right)
\end{align}

\begin{align}\notag
u_{2,j}v_{2,j} =& \left(\sum_{i=1}^D x_i^2r_{ij}\right)\left(\sum_{i=1}^D y_i^2r_{ij}\right)
=\sum_{i=1}^D x_i^2y_i^2r_{ij}^2 + \sum_{i\neq i^\prime} x_i^2r_{ij} y_{i^\prime}^2r_{i^\prime j}
\end{align}}
Thus
{\small\begin{align}\notag
\text{E}\left(u_{2,j}v_{2,j}\right) = \sum_{i=1}^D x_i^2y_i^2.
\end{align}}
Similarly, we can show
{\small\begin{align}\notag
\text{E}\left(u_{3,j}v_{1,j}\right) = \sum_{i=1}^D x_i^3y_i, \hspace{0.5in} \text{E}\left(u_{1,j}v_{3,j}\right) = \sum_{i=1}^D x_iy_i^3.
\end{align}}
Therefore,
{\scriptsize\begin{align}\notag
\text{E}\left(\hat{d}_{(4)}\right)
=&\sum_{i=1}^D x_i^4 + \sum_{i=1}^D y_i^4 +
\frac{1}{k}\left(
\sum_{j=1}^k \text{E}\left(6u_{2,j}v_{2,j} - 4u_{3,j} v_{1,j} - 4u_{1,j} v_{3,j}\right)
\right)\\\notag
=&\sum_{i=1}^D x_i^4 + \sum_{i=1}^D y_i^4+\frac{1}{k}\left(
\sum_{j=1}^k \left(6\sum_{i=1}^Dx_i^2y_i^2 - 4\sum_{i=1}^Dx_i^3y_i - 4\sum_{i=1}^Dx_iy_i^3 \right)
\right)
=d_{(4)}.
\end{align}}
To derive the variance, we need to analyze the expectation
{\small\begin{align}\notag
&\left( 6u_{2,j}v_{2,j} - 4u_{3,j} v_{1,j} - 4u_{1,j} v_{3,j}\right)^2\\\notag
=&36u_{2,j}^2v_{2,j}^2 + 16u_{3,j}^2 v_{1,j}^2 +16u_{1,j}^2 v_{3,j}^2- 48u_{2,j}u_{3,j}v_{2,j}v_{1,j}\\\notag
 &\hspace{0.1in} - 48u_{2,j}u_{1,j}v_{2,j}v_{3,j} +32u_{3,j}u_{1,j}v_{1,j}v_{3,j}.
\end{align}}
To simplify the expression, we will skip the terms that will be zeros when taking expectations.
{\small\begin{align}\notag
\text{E}\left(u_{2,j}^2v_{2,j}^2\right)
=&\text{E}\left(\left(\sum_{i=1}^D x_i^2y_i^2r_{ij}^2 + \sum_{i\neq i^\prime} x_i^2r_{ij} y_{i^\prime}^2r_{i^\prime j}\right)^2\right)\\\notag
=&\text{E}\left(\sum_{i=1}^D x_i^4y_i^4r_{ij}^4 +2\sum_{i\neq i^\prime} x_i^2y_{i}^2r_{ij}^2 x_{i^\prime}^2y_{i^\prime}^2r_{i^\prime j}^2+ \sum_{i\neq i^\prime} x_i^4r_{ij}^2 y_{i^\prime}^4r_{i^\prime j}^2\right)\\\notag
=&\sum_{i=1}^D 3x_i^4y_i^4 +2\sum_{i\neq i^\prime} x_i^2y_{i}^2 x_{i^\prime}^2y_{i^\prime}^2+ \sum_{i\neq i^\prime} x_i^4 y_{i^\prime}^4\\\notag
=&\sum_{i=1}^Dx_i^4\sum_{i=1}^Dy_i^4 +  2\left(\sum_{i=1}^Dx_i^2y_i^2\right)^2.
\end{align}}
Similarly
{\small\begin{align}\notag
&\text{E}\left(u_{3,j}^2v_{1,j}^2\right)
=\sum_{i=1}^Dx_i^6\sum_{i=1}^Dy_i^2 +  2\left(\sum_{i=1}^Dx_i^3y_i\right)^2,\\\notag
&\text{E}\left(u_{3,j}^2v_{1,j}^2\right)
=\sum_{i=1}^Dx_i^2\sum_{i=1}^Dy_i^6 +  2\left(\sum_{i=1}^Dx_iy_i^3\right)^2.
\end{align}}

{\small\begin{align}\notag
&\text{E}\left(u_{2,j}u_{3,j}v_{2,j}v_{1,j}\right)\\\notag
=&\text{E}\left( \sum_{i=1}^D x_i^2r_{ij}\sum_{i=1}^D x_i^3r_{ij} \sum_{i=1}^D y_i^2r_{ij}\sum_{i=1}^D y_ir_{ij} \right)\\\notag
=&\text{E}\left(\left(\sum_{i=1}^D x_i^5r_{ij}^2 + \sum_{i\neq i^\prime} x_i^2r_{ij} x_{i^\prime}^3r_{i^\prime j}\right) \left(\sum_{i=1}^D y_i^3r_{ij}^2 + \sum_{i\neq i^\prime} y_i^2r_{ij} y_{i^\prime}r_{i^\prime j}\right)\right)\\\notag
=&\text{E}\left(\sum_{i=1}^D x_i^5y_i^3r_{ij}^4 + \sum_{i\neq i^\prime} x_i^5r_{ij}^2 y_{i^\prime}^3r_{i^\prime j}^2\right)\\\notag
+&\text{E}\left(\sum_{i\neq i^\prime} x_i^2y_i^2r_{ij}^2 x_{i^\prime}^3y_{i^\prime}r_{i^\prime j}^2
+\sum_{i\neq i^\prime} x_i^2y_ir_{ij}^2 x_{i^\prime}^3y_{i^\prime}^2r_{i^\prime j}^2
\right)\\\notag
=&3\sum_{i=1}^D x_i^5y_i^3 + \sum_{i\neq i^\prime} x_i^5y_{i^\prime}^3+\sum_{i\neq i^\prime} x_i^2y_i^2 x_{i^\prime}^3y_{i^\prime} + \sum_{i\neq i^\prime} x_i^2y_i x_{i^\prime}^3y_{i^\prime}^2\\\notag
=&\sum_{i=1}^D x_i^5\sum_{i=1}^Dy_i^3 +\sum_{i=1}^D x_i^2y_i^2 \sum_{i=1}^Dx_i^3y_i + \sum_{i=1}^D x_i^2y_i \sum_{i=1}^Dx_i^3y_i^2.
\end{align}}
Similarly
{\small\begin{align}\notag
&\text{E}\left(u_{2,j}u_{1,j}v_{2,j}v_{3,j}\right)\\\notag
=&\sum_{i=1}^D x_i^3\sum_{i=1}^Dy_i^5 +\sum_{i=1}^D x_iy_i^3 \sum_{i=1}^Dx_i^2y_i^2+\sum_{i=1}^D x_iy_i^2 \sum_{i=1}^Dx_i^2y_i^3,\\\notag
&\text{E}\left(u_{3,j}u_{1,j}v_{1,j}v_{3,j}\right)\\\notag
=&\sum_{i=1}^D x_i^4\sum_{i=1}^Dy_i^4 +\sum_{i=1}^D x_iy_i^3 \sum_{i=1}^Dx_i^3y_i+\sum_{i=1}^D x_iy_i \sum_{i=1}^Dx_i^3y_i^3.
\end{align}}
Therefore,
{\small\begin{align}\notag
&\text{Var}\left( 6u_{2,j}v_{2,j} - 4u_{3,j} v_{1,j} - 4u_{1,j} v_{3,j}\right) \\\notag
=&36\sum_{i=1}^Dx_i^4\sum_{i=1}^Dy_i^4 +  72\left(\sum_{i=1}^Dx_i^2y_i^2\right)^2\\\notag
+&16\sum_{i=1}^Dx_i^6\sum_{i=1}^Dy_i^2 +  32\left(\sum_{i=1}^Dx_i^3y_i\right)^2\\\notag
+&16\sum_{i=1}^Dx_i^2\sum_{i=1}^Dy_i^6 +  32\left(\sum_{i=1}^Dx_iy_i^3\right)^2\\\notag
-&48\left(\sum_{i=1}^D x_i^5\sum_{i=1}^Dy_i^3 +\sum_{i=1}^D x_i^2y_i^2 \sum_{i=1}^Dx_i^3y_i +\sum_{i=1}^D x_i^2y_i \sum_{i=1}^Dx_i^3y_i^2\right)\\\notag
-&48\left(\sum_{i=1}^D x_i^3\sum_{i=1}^Dy_i^5 +\sum_{i=1}^D x_iy_i^3 \sum_{i=1}^Dx_i^2y_i^2 +\sum_{i=1}^D x_iy_i^2 \sum_{i=1}^Dx_i^2y_i^3\right)\\\notag
+&32\left(\sum_{i=1}^D x_i^4\sum_{i=1}^Dy_i^4 +\sum_{i=1}^D x_iy_i^3 \sum_{i=1}^Dx_i^3y_i +\sum_{i=1}^D x_iy_i \sum_{i=1}^Dx_i^3y_i^3\right)\\\notag
&-\left(6\sum_{i=1}^Dx_i^2y_i^2 - 4\sum_{i=1}^Dx_i^3y_i - 4\sum_{i=1}^Dx_iy_i^3\right)^2
\end{align}}
\noindent from which it follows that
{\small\begin{align}\notag
\text{Var}\left( \hat{d}_{(4)}\right)
=&\frac{36}{k}\left(\sum_{i=1}^Dx_i^4\sum_{i=1}^Dy_i^4 +  \left(\sum_{i=1}^Dx_i^2y_i^2\right)^2\right)\\\notag
+& \frac{16}{k}\left(\sum_{i=1}^Dx_i^6\sum_{i=1}^Dy_i^2 +  \left(\sum_{i=1}^Dx_i^3y_i\right)^2\right)\\\notag
+&\frac{16}{k}\left(\sum_{i=1}^Dx_i^2\sum_{i=1}^Dy_i^6 +  \left(\sum_{i=1}^Dx_iy_i^3\right)^2\right)\\\notag
-&\frac{48}{k}\left(\sum_{i=1}^D x_i^5\sum_{i=1}^Dy_i^3  +\sum_{i=1}^D x_i^2y_i \sum_{i=1}^Dx_i^3y_i^2\right)\\\notag
-&\frac{48}{k}\left(\sum_{i=1}^D x_i^3\sum_{i=1}^Dy_i^5 +\sum_{i=1}^D x_iy_i^2 \sum_{i=1}^Dx_i^2y_i^3\right)\\\notag
+&\frac{32}{k}\left(\sum_{i=1}^D x_i^4\sum_{i=1}^Dy_i^4 +\sum_{i=1}^D x_iy_i \sum_{i=1}^Dx_i^3y_i^3\right)\\\notag
\end{align}}

\section{Proof of Lemma 3}\label{proof_lem_var_diff}

It suffices to show that
\begin{align}\notag
&\left(\sum_{i=1}^D x_i^5\sum_{i=1}^Dy_i^3  +\sum_{i=1}^D x_i^2y_i \sum_{i=1}^Dx_i^3y_i^2\right)\\\notag
+&\left(\sum_{i=1}^D x_i^3\sum_{i=1}^Dy_i^5 +\sum_{i=1}^D x_iy_i^2 \sum_{i=1}^Dx_i^2y_i^3\right)\\\notag
-&\left(\sum_{i=1}^D x_i^4\sum_{i=1}^Dy_i^4 +\sum_{i=1}^D x_iy_i \sum_{i=1}^Dx_i^3y_i^3\right)\geq 0.
\end{align}

We need to use the  arithmetic-geometric mean inequality:
\begin{align}\notag
\sum_{i=1}^n w_i  \geq n \left(\prod_{i=1}^n w_i\right)^{1/n}, \hspace{0.3in} \text{provided} \  \ w_i\geq 0.
\end{align}

Because
\begin{align}\notag
x_i^5 y_j ^3 +  x_i^3  y_j ^5 \geq 2\sqrt{
 x_i^8 y_j ^8} = 2x_i^4y_j^4 ,
\end{align}
\begin{align}\notag
&\sum_{i=1}^D x_i^5\sum_{i=1}^Dy_i^3  +\sum_{i=1}^D x_i^3\sum_{i=1}^Dy_i^5
-\sum_{i=1}^D x_i^4\sum_{i=1}^Dy_i^4 \geq 0.
\end{align}

Thus it only remains to show that
\begin{align}\notag
&\sum_{i=1}^D x_i^2y_i \sum_{i=1}^Dx_i^3y_i^2 + \sum_{i=1}^D x_iy_i^2 \sum_{i=1}^Dx_i^2y_i^3 - \sum_{i=1}^D x_iy_i \sum_{i=1}^Dx_i^3y_i^3 \geq 0,
\end{align}
for which it suffices to show that
\begin{align}\notag
&2\sum_{i=1}^D x_i^{3/2}y_i^{3/2} \sum_{i=1}^Dx_i^{5/2}y_i^{5/2} - \sum_{i=1}^D x_iy_i \sum_{i=1}^Dx_i^3y_i^3 \geq 0,
\end{align}
\noindent or equivalently, to show that, if $z_i\geq 0$ $\forall i\in[1,D]$, then
\begin{align}\label{eqn_ineq}
&f(z_i, i = 1, 2, ..., D) = 2\sum_{i=1}^D z_i^3 \sum_{i=1}^Dz_i^5 - \sum_{i=1}^D z_i^2\sum_{i=1}^Dz_i^6\geq 0.
\end{align}

Obviously, (\ref{eqn_ineq}) holds for $D=1$ and $D=2$.  To see that it is true for $D>2$, we notice that only at $(z_1 = 0, z_2 = 0, ..., z_D = 0)$, the first derivative of $f(z_i)$ is zero. We can also check that  $f(z_i = 1, i = 1, 2, ..., D)>0$. Since $f(z_i)$ is a continuous function, we know $f(z_i)\geq0$ must hold if $z_i>0$ for all $i$. There is no need to worry about the boundary case that $z_j=0$ and $z_i\geq 0$ because it is reduced to a small problem with $D^\prime = D-1$ and we have already shown the base case when $D = 1$ and $D=2$. Thus, we complete the proof.

\section{Proof of Lemma 5}\label{proof_lem_var_6}
\begin{align}\notag
&{d}_{(6)} = \sum_{i=1}^D x_i^6 + \sum_{i=1}^D y_i^6 - 20\sum_{i=1}^Dx_i^3y_i^3\\\notag
&+15\sum_{i=1}^Dx_i^2y_i^4 + 15\sum_{i=1}^Dx_i^4y_i^2 - 6\sum_{i=1}^Dx_i^5y_i - 6\sum_{i=1}^Dx_iy_i^5
\end{align}

\begin{align}\notag
&\hat{d}_{(6)} = \sum_{i=1}^D x_i^6 + \sum_{i=1}^D y_i^6+ \frac{1}{k}\left(-20u_3^\text{T}v_3 + 15u_4^\text{T}v_2+ 15u_2^\text{T}v_4-6u_5^\text{T}v_3- 6u_1^\text{T}v_5\right)\\\notag
&=\sum_{i=1}^D x_i^6 + \sum_{i=1}^D y_i^6 + \frac{1}{k}
\sum_{j=1}^k -20u_{3,j}v_{3,j} + 15u_{2,j} v_{4,j} + 15u_{4,j} v_{2,j} - 6u_{1,j} v_{5,j} - 6u_{5,j} v_{1,j}.
\end{align}

To derive the variance, we need to analyze the expectation of
\begin{align}\notag
&\left( -20u_{3,j}v_{3,j} + 15u_{2,j} v_{4,j} + 15u_{4,j} v_{2,j} - 6u_{1,j} v_{5,j} - 6u_{5,j} v_{1,j}\right)^2\\\notag
&=400u_{3,j}^2v_{3,j}^2 +  225u_{2,j}^2 v_{4,j}^2+225u_{4,j}^2 v_{2,j}^2 +36u_{1,j}^2 v_{5,j}^2 + 36u_{5,j}^2 v_{1,j}^2\\\notag
&-600u_{3,j}v_{3,j}u_{2,j} v_{4,j} - 600u_{3,j}v_{3,j}u_{4,j} v_{2,j} + 240u_{3,j}v_{3,j}u_{1,j} v_{5,j}\\\notag
&+240u_{3,j}v_{3,j}u_{5,j} v_{1,j}+450u_{2,j} v_{4,j}u_{4,j} v_{2,j} -180u_{2,j} v_{4,j}u_{1,j} v_{5,j}\\\notag
&-180u_{2,j} v_{4,j}u_{5,j} v_{1,j}-180u_{4,j} v_{2,j}u_{1,j} v_{5,j}-180u_{4,j} v_{2,j}u_{5,j} v_{1,j}\\\notag
&+72u_{1,j} v_{5,j}u_{5,j} v_{1,j}
\end{align}
Skipping the detail, we can show that
\begin{align}\notag
&\text{E}\left(u_{3,j}^2v_{3,j}^2\right)
=\sum_{i=1}^Dx_i^6\sum_{i=1}^Dy_i^6 +  2\left(\sum_{i=1}^Dx_i^3y_i^3\right)^2,\\\notag
&\text{E}\left(u_{2,j}^2v_{4,j}^2\right)
=\sum_{i=1}^Dx_i^4\sum_{i=1}^Dy_i^8 +  2\left(\sum_{i=1}^Dx_i^2y_i^4\right)^2,\\\notag
&\text{E}\left(u_{4,j}^2v_{2,j}^2\right)
=\sum_{i=1}^Dx_i^8\sum_{i=1}^Dy_i^4 +  2\left(\sum_{i=1}^Dx_i^4y_i^2\right)^2,\\\notag
&\text{E}\left(u_{1,j}^2v_{5,j}^2\right)
=\sum_{i=1}^Dx_i^2\sum_{i=1}^Dy_i^{10} +  2\left(\sum_{i=1}^Dx_iy_i^5\right)^2,\\\notag
&\text{E}\left(u_{5,j}^2v_{1,j}^2\right)
=\sum_{i=1}^Dx_i^{10}\sum_{i=1}^Dy_i^2 +  2\left(\sum_{i=1}^Dx_i^5y_i\right)^2.
\end{align}
And
\begin{align}\notag
&\text{E}\left(u_{3,j}u_{2,j}v_{3,j}v_{4,j}\right)\\\notag
=&\sum_{i=1}^D x_i^5\sum_{i=1}^Dy_i^7 +\sum_{i=1}^D x_i^3y_i^3 \sum_{i=1}^Dx_i^2y_i^4 + \sum_{i=1}^D x_i^3y_i^4 \sum_{i=1}^Dx_i^2y_i^3,\\\notag
&\text{E}\left(u_{3,j}u_{4,j}v_{3,j}v_{2,j}\right)\\\notag
=&\sum_{i=1}^D x_i^7\sum_{i=1}^Dy_i^5 +\sum_{i=1}^D x_i^3y_i^3 \sum_{i=1}^Dx_i^4y_i^2+\sum_{i=1}^D x_i^3y_i^2 \sum_{i=1}^Dx_i^4y_i^3,\\\notag
&\text{E}\left(u_{3,j}u_{1,j}v_{3,j}v_{5,j}\right)\\\notag
=&\sum_{i=1}^D x_i^4\sum_{i=1}^Dy_i^8 +\sum_{i=1}^D x_i^3y_i^3 \sum_{i=1}^Dx_iy_i^5+\sum_{i=1}^D x_i^3y_i^5 \sum_{i=1}^Dx_iy_i^3,\\\notag
&\text{E}\left(u_{3,j}u_{5,j}v_{3,j}v_{1,j}\right)\\\notag
=&\sum_{i=1}^D x_i^8\sum_{i=1}^Dy_i^4 +\sum_{i=1}^D x_i^3y_i^3 \sum_{i=1}^Dx_i^5y_i^1 + \sum_{i=1}^D x_i^3y_i \sum_{i=1}^Dx_i^5y_i^3,\\\notag
&\text{E}\left(u_{2,j}u_{4,j}v_{4,j}v_{2,j}\right)\\\notag
=&\sum_{i=1}^D x_i^6\sum_{i=1}^Dy_i^6 +\sum_{i=1}^D x_i^2y_i^4 \sum_{i=1}^Dx_i^4y_i^2+\sum_{i=1}^D x_i^2y_i^2 \sum_{i=1}^Dx_i^4y_i^4,\\\notag
&\text{E}\left(u_{2,j}u_{1,j}v_{4,j}v_{5,j}\right)\\\notag
=&\sum_{i=1}^D x_i^3\sum_{i=1}^Dy_i^9 +\sum_{i=1}^D x_i^2y_i^4 \sum_{i=1}^Dx_iy_i^5+\sum_{i=1}^D x_i^2y_i^5 \sum_{i=1}^Dx_iy_i^4,\\\notag
&\text{E}\left(u_{2,j}u_{5,j}v_{4,j}v_{1,j}\right)\\\notag
=&\sum_{i=1}^D x_i^7\sum_{i=1}^Dy_i^5 +\sum_{i=1}^D x_i^2y_i^4 \sum_{i=1}^Dx_i^5y_i+\sum_{i=1}^D x_i^2y_i \sum_{i=1}^Dx_i^5y_i^4,\\\notag
&\text{E}\left(u_{4,j}u_{1,j}v_{2,j}v_{5,j}\right)\\\notag
=&\sum_{i=1}^D x_i^5\sum_{i=1}^Dy_i^7 +\sum_{i=1}^D x_i^4y_i^2 \sum_{i=1}^Dx_iy_i^5+\sum_{i=1}^D x_i^4y_i^5 \sum_{i=1}^Dx_iy_i^2,\\\notag
&\text{E}\left(u_{4,j}u_{5,j}v_{2,j}v_{1,j}\right)\\\notag
=&\sum_{i=1}^D x_i^9\sum_{i=1}^Dy_i^3 +\sum_{i=1}^D x_i^4y_i^2 \sum_{i=1}^Dx_i^5y_i+\sum_{i=1}^D x_i^4y_i \sum_{i=1}^Dx_i^5y_i^2,\\\notag
&\text{E}\left(u_{1,j}u_{5,j}v_{5,j}v_{1,j}\right)\\\notag
=&\sum_{i=1}^D x_i^6\sum_{i=1}^Dy_i^6 +\sum_{i=1}^D x_iy_i^5 \sum_{i=1}^Dx_i^5y_i+\sum_{i=1}^D x_iy_i \sum_{i=1}^Dx_i^5y_i^5,
\end{align}

Combining the results, we obtain
\begin{align}\notag
\text{Var}\left( \hat{d}_{(6)}\right)
&=\frac{400}{k}\left(\sum_{i=1}^Dx_i^6\sum_{i=1}^Dy_i^6 +  \left(\sum_{i=1}^Dx_i^3y_i^3\right)^2\right)\\\notag
&+\frac{225}{k}\left(\sum_{i=1}^Dx_i^4\sum_{i=1}^Dy_i^8 +  \left(\sum_{i=1}^Dx_i^2y_i^4\right)^2\right)\\\notag
&+\frac{225}{k}\left(\sum_{i=1}^Dx_i^8\sum_{i=1}^Dy_i^4 +  \left(\sum_{i=1}^Dx_i^4y_i^2\right)^2\right)\\\notag
&+\frac{36}{k}\left(\sum_{i=1}^Dx_i^2\sum_{i=1}^Dy_i^{10} +  \left(\sum_{i=1}^Dx_iy_i^5\right)^2\right)\\\notag
&+\frac{36}{k}\left(\sum_{i=1}^Dx_i^{10}\sum_{i=1}^Dy_i^2 +  \left(\sum_{i=1}^Dx_i^5y_i\right)^2\right)+\Delta_6
\end{align}
where
\begin{align}\notag
k\Delta_6/6=&-100\left(\sum_{i=1}^D x_i^5\sum_{i=1}^Dy_i^7 + \sum_{i=1}^D x_i^3y_i^4 \sum_{i=1}^Dx_i^2y_i^3\right)\\\notag
&-100\left(\sum_{i=1}^D x_i^7\sum_{i=1}^Dy_i^5 +\sum_{i=1}^D x_i^3y_i^2 \sum_{i=1}^Dx_i^4y_i^3\right)\\\notag
&+40\left(\sum_{i=1}^D x_i^4\sum_{i=1}^Dy_i^8 +\sum_{i=1}^D x_i^3y_i^5 \sum_{i=1}^Dx_iy_i^3\right)\\\notag
&+40\left(\sum_{i=1}^D x_i^8\sum_{i=1}^Dy_i^4 + \sum_{i=1}^D x_i^3y_i \sum_{i=1}^Dx_i^5y_i^3\right)\\\notag
&+75\left(\sum_{i=1}^D x_i^6\sum_{i=1}^Dy_i^6 +\sum_{i=1}^D x_i^2y_i^2 \sum_{i=1}^Dx_i^4y_i^4\right)\\\notag
&-30\left(\sum_{i=1}^D x_i^3\sum_{i=1}^Dy_i^9 +\sum_{i=1}^D x_i^2y_i^5 \sum_{i=1}^Dx_iy_i^4\right)\\\notag
&-30\left(\sum_{i=1}^D x_i^7\sum_{i=1}^Dy_i^5 +\sum_{i=1}^D x_i^2y_i \sum_{i=1}^Dx_i^5y_i^4\right)\\\notag
&-30\left(\sum_{i=1}^D x_i^5\sum_{i=1}^Dy_i^7 +\sum_{i=1}^D x_i^4y_i^5 \sum_{i=1}^Dx_iy_i^2\right)\\\notag
&-30\left(\sum_{i=1}^D x_i^9\sum_{i=1}^Dy_i^3 +\sum_{i=1}^D x_i^4y_i \sum_{i=1}^Dx_i^5y_i^2\right)\\\notag
&+12\left(\sum_{i=1}^D x_i^6\sum_{i=1}^Dy_i^6 +\sum_{i=1}^D x_iy_i \sum_{i=1}^Dx_i^5y_i^5\right).
\end{align}

\section{Proof of Lemma 6}\label{proof_lem_var_subG}

\begin{align}\notag
&\hat{d}_{(4),s} = \sum_{i=1}^D x_i^4 + \sum_{i=1}^D y_i^4 + \frac{1}{k}\left(6u_2^\text{T}v_2 - 4u_3^\text{T}v_1- 4u_1^\text{T}v_3\right)\\\notag
=&\sum_{i=1}^D x_i^4 + \sum_{i=1}^D y_i^4 + \frac{1}{k}\left(
\sum_{j=1}^k 6u_{2,j}v_{2,j} - 4u_{3,j} v_{1,j} - 4u_{1,j} v_{3,j}
\right)
\end{align}

\begin{align}\notag
\text{E}\left(u_{2,j}^2v_{2,j}^2\right)
=&\text{E}\left(\left(\sum_{i=1}^D x_i^2y_i^2r_{ij}^2 + \sum_{i\neq i^\prime} x_i^2r_{ij} y_{i^\prime}^2r_{i^\prime j}\right)^2\right)\\\notag
=&\text{E}\left(\sum_{i=1}^D x_i^4y_i^4r_{ij}^4 +2\sum_{i\neq i^\prime} x_i^2y_{i}^2r_{ij}^2 x_{i^\prime}^2y_{i^\prime}^2r_{i^\prime j}^2+ \sum_{i\neq i^\prime} x_i^4r_{ij}^2 y_{i^\prime}^4r_{i^\prime j}^2\right)\\\notag
=&\sum_{i=1}^D s \ x_i^4y_i^4 +2\sum_{i\neq i^\prime} x_i^2y_{i}^2 x_{i^\prime}^2y_{i^\prime}^2+ \sum_{i\neq i^\prime} x_i^4 y_{i^\prime}^4\\\notag
=&\sum_{i=1}^Dx_i^4\sum_{i=1}^Dy_i^4 +  2\left(\sum_{i=1}^Dx_i^2y_i^2\right)^2 + (s-3)\sum_{i=1}^D x_i^4y_i^4.
\end{align}
Similarly,
\begin{align}\notag
&\text{E}\left(u_{3,j}^2v_{1,j}^2\right)
=\sum_{i=1}^Dx_i^6\sum_{i=1}^Dy_i^2 +  2\left(\sum_{i=1}^Dx_i^3y_i\right)^2 + (s-3)\sum_{i=1}^D x_i^6y_i^2,\\\notag
&\text{E}\left(u_{3,j}^2v_{1,j}^2\right)
=\sum_{i=1}^Dx_i^2\sum_{i=1}^Dy_i^6 +  2\left(\sum_{i=1}^Dx_iy_i^3\right)^2+(s-3)\sum_{i=1}^D x_i^2y_i^6.
\end{align}
\begin{align}\notag
&\text{E}\left(u_{2,j}u_{3,j}v_{2,j}v_{1,j}\right) =\sum_{i=1}^D x_i^5\sum_{i=1}^Dy_i^3 +\sum_{i=1}^D x_i^2y_i^2 \sum_{i=1}^Dx_i^3y_i\\\notag
&\hspace{0.8in} + \sum_{i=1}^D x_i^2y_i \sum_{i=1}^Dx_i^3y_i^2+(s-3)\sum_{i=1}^D x_i^5y_i^3,\\\notag
&\text{E}\left(u_{2,j}u_{1,j}v_{2,j}v_{3,j}\right)=\sum_{i=1}^D x_i^3\sum_{i=1}^Dy_i^5 +\sum_{i=1}^D x_iy_i^3 \sum_{i=1}^Dx_i^2y_i^2\\\notag
&\hspace{0.8in} +\sum_{i=1}^D x_iy_i^2 \sum_{i=1}^Dx_i^2y_i^3+(s-3)\sum_{i=1}^D x_i^3y_i^5,\\\notag
&\text{E}\left(u_{3,j}u_{1,j}v_{1,j}v_{3,j}\right)=\sum_{i=1}^D x_i^4\sum_{i=1}^Dy_i^4 +\sum_{i=1}^D x_iy_i^3 \sum_{i=1}^Dx_i^3y_i\\\notag
&\hspace{0.8in}+\sum_{i=1}^D x_iy_i \sum_{i=1}^Dx_i^3y_i^3+(s-3)\sum_{i=1}^D x_i^4y_i^4.
\end{align}
Therefore,
{\scriptsize
\begin{align}\notag
&\text{Var}\left( 6u_{2,j}v_{2,j} - 4u_{3,j} v_{1,j} - 4u_{1,j} v_{3,j}\right) \\\notag
=&36\sum_{i=1}^Dx_i^4\sum_{i=1}^Dy_i^4 +  72\left(\sum_{i=1}^Dx_i^2y_i^2\right)^2+36(s-3)\sum_{i=1}^D x_i^4y_i^4\\\notag
+& 16\sum_{i=1}^Dx_i^6\sum_{i=1}^Dy_i^2 +  32\left(\sum_{i=1}^Dx_i^3y_i\right)^2+36(s-3)\sum_{i=1}^D x_i^6y_i^2\\\notag
+&16\sum_{i=1}^Dx_i^2\sum_{i=1}^Dy_i^6 +  32\left(\sum_{i=1}^Dx_iy_i^3\right)^2+36(s-3)\sum_{i=1}^D x_i^2y_i^6\\\notag
-&48\left(\sum_{i=1}^D x_i^5\sum_{i=1}^Dy_i^3 +\sum_{i=1}^D x_i^2y_i^2 \sum_{i=1}^Dx_i^3y_i +\sum_{i=1}^D x_i^2y_i \sum_{i=1}^Dx_i^3y_i^2+(s-3)\sum_{i=1}^D x_i^5y_i^3\right)\\\notag
-&48\left(\sum_{i=1}^D x_i^3\sum_{i=1}^Dy_i^5 +\sum_{i=1}^D x_iy_i^3 \sum_{i=1}^Dx_i^2y_i^2 +\sum_{i=1}^D x_iy_i^2 \sum_{i=1}^Dx_i^2y_i^3+(s-3)\sum_{i=1}^D x_i^3y_i^5\right)\\\notag
+&32\left(\sum_{i=1}^D x_i^4\sum_{i=1}^Dy_i^4 +\sum_{i=1}^D x_iy_i^3 \sum_{i=1}^Dx_i^3y_i +\sum_{i=1}^D x_iy_i \sum_{i=1}^Dx_i^3y_i^3+(s-3)\sum_{i=1}^D x_i^4y_i^4\right)\\\notag
-&\left(6\sum_{i=1}^Dx_i^2y_i^2 - 4\sum_{i=1}^Dx_i^3y_i - 4\sum_{i=1}^Dx_iy_i^3\right)^2
\end{align}}
\noindent from which it follows that
\begin{align}\notag
\text{Var}\left( \hat{d}_{(4),s}\right)
=&\frac{36}{k}\left(\sum_{i=1}^Dx_i^4\sum_{i=1}^Dy_i^4 +  \left(\sum_{i=1}^Dx_i^2y_i^2\right)^2+(s-3)\sum_{i=1}^D x_i^4y_i^4\right)\\\notag
+& \frac{16}{k}\left(\sum_{i=1}^Dx_i^6\sum_{i=1}^Dy_i^2 +  \left(\sum_{i=1}^Dx_i^3y_i\right)^2+(s-3)\sum_{i=1}^D x_i^6y_i^2\right)\\\notag
+&\frac{16}{k}\left(\sum_{i=1}^Dx_i^2\sum_{i=1}^Dy_i^6 +  \left(\sum_{i=1}^Dx_iy_i^3\right)^2+(s-3)\sum_{i=1}^D x_i^2y_i^6\right)\\\notag
-&\frac{48}{k}\left(\sum_{i=1}^D x_i^5\sum_{i=1}^Dy_i^3  +\sum_{i=1}^D x_i^2y_i \sum_{i=1}^Dx_i^3y_i^2+(s-3)\sum_{i=1}^D x_i^5y_i^3\right)\\\notag
-&\frac{48}{k}\left(\sum_{i=1}^D x_i^3\sum_{i=1}^Dy_i^5 +\sum_{i=1}^D x_iy_i^2 \sum_{i=1}^Dx_i^2y_i^3+(s-3)\sum_{i=1}^D x_i^3y_i^5\right)\\\notag
+&\frac{32}{k}\left(\sum_{i=1}^D x_i^4\sum_{i=1}^Dy_i^4 +\sum_{i=1}^D x_iy_i \sum_{i=1}^Dx_i^3y_i^3+(s-3)\sum_{i=1}^D x_i^4y_i^4\right)\\\notag
\end{align}

\end{document}